\documentclass{article}
\usepackage[numbers]{natbib}

\usepackage{graphicx,tabularx}
\usepackage{subfigure}
\usepackage{booktabs} 
\usepackage{amsmath,amsfonts,amsthm,graphicx,float,color,tikz,url}
\usepackage{mathtools}

\usepackage{hyperref}

\newtheorem{definition}{Definition}

\newtheorem{theorem}{Theorem}
\newtheorem{proposition}{Proposition} 

\newtheorem{example}{Example}

\usepackage[nonatbib,preprint]{neurips_2020}
\usepackage[utf8]{inputenc} 
\usepackage[T1]{fontenc}    
\usepackage{hyperref}       
\usepackage{url}            
\usepackage{booktabs}       
\usepackage{amsfonts}       
\usepackage{nicefrac}       
\usepackage{microtype}      

\title{Fair Classification via Unconstrained Optimization}

\author{%
  Ibrahim M.~Alabdulmohsin \\
  Google Research, Brain Team\\
  Z\"urich, Switzerland \\
  \texttt{ibomohsin@google.com} 
}

\begin{document}

\maketitle

\begin{abstract}
Achieving the Bayes optimal binary classification rule subject to group fairness constraints is known to be reducible, in some cases, to learning a group-wise thresholding rule over the Bayes regressor. In this paper, we extend this result by proving that, in a broader setting, the Bayes optimal fair learning rule remains a group-wise thresholding rule over the Bayes regressor but with a (possible) randomization at the thresholds. This provides a stronger justification to the post-processing approach in fair classification, in which (1) a predictor is learned first, after which (2) its output is adjusted to remove bias. We show how the post-processing rule in this two-stage approach can be learned quite efficiently by solving an \emph{unconstrained} optimization problem. The proposed algorithm can be applied to any black-box machine learning model, such as deep neural networks, random forests and support vector machines. In addition, it can accommodate many fairness criteria that have been previously proposed in the literature, such as equalized odds and statistical parity. We prove that the algorithm is Bayes consistent and motivate it, furthermore, via an impossibility result that quantifies the tradeoff between accuracy and fairness across multiple demographic groups. Finally, we conclude by validating the algorithm on the Adult benchmark dataset. 
\end{abstract} 

\section{Introduction}
Machine learning applications are being increasingly adopted to make life-critical decisions with an ever-lasting impact on individual lives, such as for credit lending \cite{bruckner2018promise}, medical applications \cite{deo2015machine}, and criminal justice \cite{brennan2009evaluating}. Consequently, it is imperative to ensure that such automated decision-making systems abstain from ethical malpractice, including \lq\lq bias." 

Unfortunately, despite the fact that bias (or \lq\lq fairness") is a central concept in our society today, it is difficult to define it in precise terms. In fact, because people perceive ethical matters differently depending on their geographical location and culture \cite{awad2018moral}, no universally-agreeable definition for bias exists! Moreover, even within the same cultural group, bias may be defined differently according to the application at hand and might even be ignored in favor of accuracy by some members of the group when the stakes are high, such as for medical diagnosis \cite{kleinberg2016inherent,ingold2016amazon}.

As such, it is not surprising to note that several definitions for \lq\lq unbiased classifiers" have been introduced in the literature. These include statistical parity \cite{dwork2012fairness,zafar2017fairness}, equality of opportunity \cite{hardt2016equality}, and equalized odds \cite{kleinberg2016inherent}. Unfortunately, such definitions are not generally mutually compatible \cite{chouldechova2017fair,kleinberg2016inherent} and some might even be in conflict with calibration \cite{kleinberg2016inherent}. In addition, because fairness is a societal concept, it does not necessarily translate into a statistical criteria \cite{chouldechova2017fair,dixon2018measuring}.

In order to contrast the previous definitions of bias in the literature, let $\mathcal{X}$ be an instance space and let $\mathcal{Y}=\{0,1\}$ be the target set in a standard binary classification problem. In the fair classification setting, we may further assume the existence of a (possibly randomized) sensitive attribute $1_S:\mathcal{X}\to\{0,1\}$, where $1_S(x)$ indicates if an instance $x\in\mathcal{X}$ belongs to some sensitive class $S\subseteq\mathcal{X}$. For example, $\mathcal{X}$ might correspond to the set of job applicants while $S$ are the female applicants. Then, a general theme among many popular definitions of bias is to fix a partitioning of the instance space $\mathcal{X}=\cup_k X_k$ where $X_{k_1}\cap X_{k_2}=\emptyset$ for $k_1\neq k_2$ and require a balanced representation of the sensitive class within some or all of the subsets $X_k$. Throughout this paper, we refer to the subsets $X_k\subseteq\mathcal{X}$ as \lq\lq groups." This brings us to the first definition:

\begin{definition}[Conditional Statistical Parity]\label{def::conditional_convar}
A classifier $f:\mathcal{X}\to\{0,1\}$ is said to be conditionally unbiased in some group $X_k\subseteq\mathcal{X}$ with respect to a sensitive class $S$ if $ C\big(f(\textbf{\textbf{x}}),\,1_S(\textbf{x});\;\textbf{x}\in X_k\big) = 0$, where for any binary random variables $\textbf{a, b}\in\{0,1\}$, $C(\textbf{a}, \textbf{b}) \doteq \mathbb{E}[\textbf{a}\cdot\textbf{b}] - \mathbb{E}[\textbf{a}]\cdot \mathbb{E}[\textbf{b}]$ is their covariance, and $C(\textbf{a}, \textbf{b};\,\textbf{c})$ is their covariance  conditioned on $\textbf{c}\in\{0,1\}$:
\begin{equation}
    C(\textbf{a}, \textbf{b};\,\textbf{c}=c) = \mathbb{E}[\textbf{a}\cdot\textbf{b}\,|\,\textbf{c}=c] - \mathbb{E}[\textbf{a}\,|\,\textbf{c}=c]\cdot \mathbb{E}[\textbf{b}\,|\,\textbf{c}=c]
\end{equation}
A classifier $f$ is said to be unbiased with respect to $S$ across all groups $\{X_k\}_{k=1,2,\ldots,K}$ if it is conditionally unbiased with respect to $S$ in each $X_k$ separately.
\end{definition}

We note that statistical parity corresponds to bias in the group $\mathcal{X}$, equality of opportunity corresponds to bias in the group $X = \{x\in\mathcal{X}:\,y(x)=1\}$, while equalized odds corresponds to bias in both groups $X$ and $\mathcal{X}\setminus X$ simultaneously, where $X$ is as defined previously. Obviously, countless other possibilities exist depending on the choices of $X_k$. 


Often, machine learning algorithms are required to treat multiple groups equally. For example, the US Equal Credit Opportunity Act of 1974 \cite{usequalcredit1974} prohibits any discrimination based on gender, race, color, religion, national origin, marital status, or age. In such case, $X_k$ would correspond to a combination of multiple attributes, such as \lq\lq black Muslim females."  This brings us to the second definition:
\begin{definition}[Predictive Equality]\label{def::equal_treatment}
A classifier $f:\mathcal{X}\to\{0,1\}$ is said to satisfy predictive equality across all groups $\mathcal{G} = \{X_1,\ldots,X_k\} $ if for all $X_k\in\mathcal{G}$, one has $C\big(f(\textbf{\textbf{x}}),\,1_{X_k}(\textbf{x})\big) = 0$.
\end{definition}
Both Definitions \ref{def::conditional_convar} and \ref{def::equal_treatment} are standard in the literature (see for instance \cite{corbett2017algorithmic} and the notion of \lq\lq sub-group fairness" in \cite{mehrabi2019survey}). The difference between both definitions is that the probability of having $f(x)=1$ can vary in Definition \ref{def::conditional_convar} from one group to another as long as the sensitive class $S$ is well-represented within each group, respectively. In Definition \ref{def::equal_treatment}, the probability that $f(x)=1$ is required to be the same across all groups without designating a sensitive class $S$.  

\begin{example}
    Let $X_k$ be the set of job applicants with country of origin $k$ and $S$ be the class of females. Definition \ref{def::conditional_convar} is satisfied by an employer if for any country of origin $k$, the probability of hiring a female candidate equals the probability of hiring a male candidate conditioned on both being in $X_k$, even when some countries of origin are more preferred than others (such as citizens). Definition \ref{def::equal_treatment} is satisfied if the probability of hiring a job applicant is independent of the country of origin.
\end{example}

Under certain assumptions, it has been shown that the optimal binary classification rule subject to certain fairness constraints reduces to a group-wise thresholding rules over the Bayes regressor $\eta(x) = p(\textbf{y}=1|\textbf{x}=x)$. This happens, for example, when maximizing a social utility score that includes the cost of predicting the positive class (e.g. detaining individuals) and the positive regressor has a positive density in the unit interval \cite{corbett2017algorithmic}. In this paper, we extend this result by proving that, in a broader setting with relaxed assumptions, the Bayes optimal fair learning rule that minimizes the 0-1 misclassification loss  remains a group-wise thresholding rule over the Bayes regressor but with a (possible) randomization at the thresholds. Since this holds in a broader setting than previously considered, it provides a stronger justification to the post-processing approach in fair classification, in which (1) a predictor is learned first, after which (2) its output is adjusted to remove bias.

In addition, we show how the post-processing rule in this two-stage approach can be learned quite efficiently by, first, formulating it as an \emph{unconstrained} optimization problem, and, then, solving it using the  stochastic gradient descent (SGD) method. We argue that this approach has many distinct advantages. First, stochastic convex optimization methods are well-understood and can scale well to massive amounts of data \cite{bottou2010large}. In particular, the unconstrained loss being minimized in the proposed algorithm, which can be interpreted as a smooth approximation to the rectified linear unit (ReLU) \cite{nair2010rectified}, has many nice properties, such as Lipschitz continuity and differentiability, which imply fast convergence. Second, the guarantees provided by our algorithm hold w.r.t. the \emph{binary} predictions instead of using a proxy, such as the margin as in some previous works \cite{pmlr-v54-zafar17a,JMLR:v20:18-262}. Third, we prove that our algorithm is Bayes consistent if the original pre-processed classifier is itself Bayes consistent.

\section{Related Work}
The literature on fairness has been growing quite rapidly. This includes methods for quantifying bias using economics and social welfare theory, developing impossibility results, and proposing algorithms in both the supervised and unsupervised learning setting. In general, algorithms for fair machine learning can be broadly classified into three groups: (1) pre-processing methods, (2) in-processing methods, and (3) post-processing methods \cite{JMLR:v20:18-262}. We briefly survey them  in this section.

\textbf{Preprocessing:} The goal of preprocessing algorithms is to transform the data $D = \{(\textbf{x}_i,\textbf{y}_i)\}_{i=1,\ldots,N}$ into a different representation $\tilde D$  such that any classifier trained on $\tilde D$ in will not exhibit bias. This includes methods for learning a fair representation  \cite{pmlr-v28-zemel13,lum2016statistical,bolukbasi2016man,calmon2017optimized,pmlr-v80-madras18a,kamiran2012data} or label manipulation  \cite{kamiran2009classifying}. In some cases, such as text classification, a similar objective can be achieved by augmenting the training set with additional examples from other sources to mitigate the unintended bias of the classifier  \cite{dixon2018measuring}. Recently, \cite{locatello2019fairness} showed that unintended bias could occur even when using the Bayes optimal classifier and the target label was independent of the sensitive class. The intuition behind this result is that even if the target $\textbf{y}$ is independent of the sensitive class, it can be conditionally dependent on it given $\textbf{x}$. They further showed that \emph{disentangled} representation could mitigate this effect. 

\textbf{In-Processing:} In-processing methods constrain the behavior of learning algorithms in order to control bias. This includes methods based on adversarial learning \cite{zhang2018mitigating}, which adjust the gradient updates, and constraint-based classification, such as by incorporating constrains on the decision margin of the classifiers \cite{JMLR:v20:18-262} or on the choice of features \cite{grgic2018beyond}. Importantly, \cite{pmlr-v80-agarwal18a} showed that the problem of learning an unbiased classifier could be modeled as a cost-sensitive classification problem, which, in turn, could be applied to any \emph{black-box} classifier. One caveat of the latter approach is that it required solving a linear program (LP) and training classifiers \emph{several} times until convergence. 


\textbf{Post-Processing:} The algorithm we propose in this paper is a post-processing method. The analysis of \cite{corbett2017algorithmic} provides one theoretical justification for our approach by showing that the optimal decision rule that the maximizes social utility subject to fairness constraints is a group-dependent thresholding rule on the Bayes regressor. In \cite{hardt2016equality}, a method is proposed for postprocessing the output of a classifier using linear programming (LP) to satisfy equalized odds or equality of opportunity. 

\textbf{Definitions of Bias:} As mentioned earlier, many definitions of bias or fairness have been proposed in the literature. On one hand, there is \emph{group fairness}, such as statistical parity \cite{dwork2012fairness,mehrabi2019survey}, equalized odds \cite{kleinberg2016inherent}, and equality of opportunity/disparate mistreatment \cite{zafar2017fairness,hardt2016equality}. On the other hand, there is \emph{individual fairness}, which postulates that similar individuals need to be treated similarly \cite{dwork2012fairness}. Unfortunately, individual fairness, while appealing, transfers the burden of defining fairness from having an appropriate statistical criteria to defining a socially acceptable measure of distance/similarity between individuals. 

\textbf{Societal Impact:} Often, the central motivation behind controlling unintended bias is that it will improve the welfare of the protected group. The work of \cite{pmlr-v80-liu18c} challenges this view by illustrating how controlling bias in the learning algorithm could actually harm the protected group over the long term. The intuition behind this result is that misclassification can inflict harm on the protected class, such as by negatively impacting their credit history. Along related lines, \cite{speicher2018unified} addresses the question of how to quantify bias in the first place. They show that classical measures of inequality in economics and social welfare were motivated by axioms that remain valid for measuring bias and, hence, can be adopted for this particular purpose. In our case, we assume that fairness constraints are fixed (e.g. by Law) and do not discuss whether or not such constraints are justifiable. 

\textbf{Impossibility Results:} Recent works have established several impossibility results related to fair classification. For example,  \cite{kleinberg2016inherent,chouldechova2017fair} showed that statistical parity and equalized odds may not be achieved simultaneously under certain settings. Also, \cite{kleinberg2016inherent} showed that equalized odds might even be in conflict with probability calibration of the classifier. In our case, we derive a new impossibility result that holds for \emph{any} deterministic binary classifier and \emph{any} sensitive attribute except under degenerate conditions, which  quantifies the tradeoff between accuracy and fairness. We use this impossibility result to motivate the development of the main algorithm in Section \ref{sect::fair_via_unconstrainted}. 

\textbf{Other Settings:} Bias can be defined in other learning settings besides classification. For instance, \cite{fitzsimons2019general} presents a method for incorporating fair constraints into kernel regression methods. Also, \cite{schmidt2018fair,kleindessner2019fair,kleindessner2019guarantees,cucuringu2016simple} present algorithms for unbiased clustering, including $k$-means and spectral methods. Moreover, unbiased methods have been proposed for dimensionality reduction \cite{samadi2018price} and online learning \cite{joseph2016fairness}, among others. In our case, we focus on the binary classification setting.  

\section{Preliminaries}
As stated earlier, the objective is to produce an unbiased classifier according to either Definition \ref{def::conditional_convar} or \ref{def::equal_treatment}. This encompasses many important fairness criteria in the literature, such as statistical parity and equalized odds. Before doing that, we show that the Bayes optimal decision rule that satisfies a wider setting of affine constraints, which contain both Definitions  \ref{def::conditional_convar} or \ref{def::equal_treatment}, is a group-wise thresholding rule with (possible) randomization at the thresholds. 

\begin{theorem}\label{prop::decision_rules}
Let $f^\star = \arg\min_{f:\mathcal{X}\to\{0,1\}}\mathbb{E}[ \mathbb{I}\{f(\textbf{x})\neq \textbf{y}\}]$ be the Bayes optimal decision rule subject to group-wise affine constraints of the form $\mathbb{E}[w_k(\textbf{x})\cdot f(\textbf{x})\,|\,\textbf{x}\in X_k] = b_k$ for some fixed partition $\mathcal{X}=\cup_k X_k$. If $w_k:\mathcal{X}\to\mathbb{R}$ and $b_k\in\mathbb{R}$ are such that there exists a constant $c\in(0,1)$ in which $p(f(x)=1)=c$ will satisfy all the affine constraints, then $f^\star$ satisfies $p(f^\star(x)=1) = \mathbb{I}\{\eta(x)> t_k\} + \tau_k\, \mathbb{I}\{\eta(x)= t_k\}$, where $\eta(x)=p(\textbf{y}=1|\textbf{x}=x)$ is the Bayes regressor, $t_k\in[0,1]$ is a threshold specific to the group $X_k\subseteq\mathcal{X}$, and $\tau_k\in[0,1]$.
\end{theorem}

Theorem \ref{prop::decision_rules} is stronger than Theorem 3.2 in \cite{corbett2017algorithmic} because it holds for arbitrary group-wise affine constraints and does not assume that $\eta(x)$ has a positive density in the unit interval. In addition, the condition on $c$ is satisfied for both Definitions \ref{def::conditional_convar} and \ref{def::equal_treatment}.  The theorem shows that randomization at the thresholds is, sometimes, necessary for achieving Bayes optimality. This is illustrated next.
\begin{example}\label{example::randomization}
Suppose that $\mathcal{X}=\{-1, 0, 1\}$ where $p(
\textbf{x}=-1)=1/2$, $p(\textbf{x}=0)=1/3$ and $p(\textbf{x}=1)=1/6$. Let $\eta(-1)=0, \,\eta(0)=1/2$ and $\eta(1)=1$. In addition, let $\textbf{s}$ be a sensitive attribute, where $p(\textbf{s}=1|\textbf{x}=-1)=1/2$, $p(\textbf{s}=1|\textbf{x}=0)=1$, and $p(\textbf{s}=1|\textbf{x}=1)=0$. Then, the Bayes optimal prediction rule $f^\star$ subject to statistical parity w.r.t. $\textbf{s}$ satisfies: $p(f^\star=1|\textbf{x}=-1)=0$, $p(f^\star=1|\textbf{x}=0)=1/2$ and $p(f^\star=1|\textbf{x}=1)=1$.  
\end{example}

One implication of Theorem \ref{prop::decision_rules} is that it suggests the following two-stage approach for learning an unbiased classifier. First, we learn a scoring function $f(x)$ that serves as an approximation to $\eta(x)$, such as using the margin in support vector machines (e.g. Platt's scaling \cite{platt1999probabilistic}) or the softmax activation output in deep neural networks (e.g. temperature scaling \cite{guo2017calibration}). Second, we adjust the output of $f$ to remove bias using a group-wise thresholding rule with randomization at the thresholds. 

Throughout our discussion, we have assumed that the groups $X_k$ in either Definition \ref{def::conditional_convar} or \ref{def::equal_treatment} are fixed. Indeed, this is necessary. Our next theorem shows that no algorithm can possibly achieve fairness across all possible groups $X_k\in 2^\mathcal{X}$ except under degenerate conditions. Hence, to circumvent such an impossibility result, one has to fix the choice of the groups $X_k$ in which bias is to be controlled. 

\begin{theorem}\label{theorem::impossibility}
Let $\mathcal{X}$ be the instance space and $\mathcal{Y}=\{0,\,1\}$ be a target set. Let $1_S:\mathcal{X}\to\{0,1\}$ be an arbitrary (possibly randomized) binary-valued function on $\mathcal{X}$ and define $\gamma:\mathcal{X}\to[0,1]$ by $\gamma(x) = p(1_S(\textbf{x})=1\,|\,\textbf{x}=x)$, 
where the probability is evaluated over the randomness of $1_S:\mathcal{X}\to\{0,1\}$. Write $\bar\gamma = \mathbb{E}_{\textbf{x}}[\gamma(\textbf{x})]$.
Then, for any binary predictor $f:\mathcal{X}\to\{0,1\}$, one has:
\begin{align}\label{eq::impossibility_theorem_main_eq}
\sup_{\pi:\, \mathcal{X}\to\{0,1\}} \Big\{\mathbb{E}_{\pi(\textbf{x})}\, \big|\mathcal{C}\big(f(\textbf{\textbf{x}}),\gamma(\textbf{x});\;\pi(\textbf{x})\big)\big|\Big\}\ge\; \frac{1}{2}\;\mathbb{E}_{\textbf{x}}|\gamma(\textbf{x})-\bar\gamma|\cdot \min\{\mathbb{E}f, 1-\mathbb{E} f\},
\end{align}
where the supremum is over all binary partitions of the instance space and $\mathcal{C}\big(f(\textbf{\textbf{x}}),\gamma(\textbf{x});\;\pi(\textbf{x})\big)$ is given by Definition \ref{def::conditional_convar}. 
\end{theorem}
A converse to Theorem \ref{theorem::impossibility} holds as well by the data processing inequality in information theory \cite{cover2012elements}. As a result, a deterministic classifier $f:\mathcal{X}\to\{0,1\}$ is universally unbiased w.r.t. a sensitive class $S$ across all possible groups $X_k\in 2^\mathcal{X}$ if and only if the representation $\textbf{x}$ carries zero mutual information with the sensitive attribute or if $f$ is constant almost everywhere. 

\section{Fair Classification via Unconstrained Optimization}\label{sect::fair_via_unconstrainted}
Next, we derive the postprocessing algorithm that coincides with the Bayes optimal approach of Theorem \ref{prop::decision_rules}. We will first focus on Definition \ref{def::conditional_convar}. After that, we describe how the algorithm can be modified to control bias according to Definition \ref{def::equal_treatment}. We analyze the convergence rate in Section \ref{sect::analysis}.

\subsection{Conditional Statistical Parity}\label{sect::postprocessing_via_qp_bias_to_sensitive_class}
Suppose we have a binary classifier on the instance space $\mathcal{X}$. We would like to construct an algorithm for post-processing the predictions made by that classifier such that we control the bias with respect to a sensitive attribute $1_S:\mathcal{X}\to\{0,1\}$ across a fixed set of pairwise disjoint groups $X_1,\ldots, X_K\subseteq\mathcal{X}$. By treating the original classifier as a black-box routine, the algorithm can be applied to any binary classification method including deep neural networks, support vector machines, and decision trees. Moreover, it can be applied to ensemble methods, such as using boosting, stacking, or bagging. 

Assume that we have $K$ pairwise disjoint groups $X_1,\ldots, X_K$ and that the output of the classifier $f:\mathcal{X}\to[-1,\,+1]$ is an estimate to $2\eta(x)-1$, where $\eta(x)$ is the Bays regressor. As mentioned earlier, many algorithms can be calibrated to provide probability scores  \cite{platt1999probabilistic, guo2017calibration} so the assumption is valid. We consider randomized rules of the form:
\begin{equation*}
    \tilde f:\mathcal{X}\times\{1,2,\ldots,K\}\times\{0,1\}\times[-1,\,1]\to\{0,1\},
\end{equation*}
whose arguments are: (1) the instance $\textbf{x}\in\mathcal{X}$, (2) the group $X_k$, (3)  the sensitive attribute $1_S({x})$, and (4) the original classifier's score $f({x})$. Because randomization is sometimes necessary as mentioned earlier, $\tilde f(x)$ is the probability of predicting the positive class when the instance is $x\in\mathcal{X}$.

Let $q_i = \tilde f(x_i)\in[0,\,1]$ for the $i$-th  training example. 
For each group $X_k\subseteq\mathcal{X}$, the fairness constraint in Definition \ref{def::conditional_convar} can be written as $(\sum_{i\in X_k\cap S} q_i)/{\sum_{i\in X_k} q_i} = \rho_k$, 
where $\rho_k = (\sum_{i\in X_k\cap S} 1)/(\sum_{i\in X_k} 1)$. Having a single constraint of this form is sufficient because:
\begin{align*}
\frac{\sum_{i\in X_k\cap S} (1-q_i)}{\sum_{i\in X_k} (1-q_i)} &= \frac{|S\cap X_k|-\sum_{i\in X_k\cap S} q_i}{|X_k|-\sum_{i\in X_k} q_i} = \frac{\rho_k |X_k|-\rho_k \sum_{i\in X_k} q_i}{|X_k|-\sum_{i\in X_k} q_i} = \rho_k
\end{align*} 
Hence, for every group $X_k$, we have a linear constraint of the form $\sum_ {i\in X_k} (1_S(i)-\rho_k)\, q_i \;=\;0$.
To learn $\tilde f$, we propose solving the following \emph{regularized} optimization problem:
\begin{align}\label{eq::final_lp}
\min_{0\le q_i\le 1} \quad\quad \sum_{i=1}^N (\gamma/2)\,q_i^2\,-\,f({x}_i)\,q_i 
    & \quad \text{s.t.} \quad\forall X_k\in\mathcal{G}: \,\sum_ {i\in X_k} (1_S(i)-\rho_k)\, q_i \;=\;0,
\end{align}
where $\gamma\ll 1$ is a regularization parameter. As will be shown later, having $\gamma>0$ will lead to a randomized decision rule near the thresholds. We establish the relation between this approach and the optimal method in Theorem \ref{prop::decision_rules} through the following series of propositions. 
\begin{proposition}\label{prop::unconstraint_opt}
    The optimization problem in (\ref{eq::final_lp}) is equivalent to the unconstrained optimization: 
    \begin{align}\label{eq::theorem_2_equiv_form}
        \min_{q_i,\mu_k\in\mathbb{R}}\sum_{X_k\in\mathcal{G}}\sum_{i\in X_k}         \Big(\frac{\gamma}{2}q_i^2 +[f({x}_i)-\gamma q_i-\mu_k(i)(1_S(i)-\rho_k)]^+\Big),
    \end{align}
    where $[x]^+ = \max\{0, \,x\}$.
\end{proposition}

We describe, next, how to use $\gamma, \mu_k$ and $\rho_k$ to adjust the predictions of the original classifier $f$. To reiterate, $1_X(x)$ is the characteristic function of the set $X$. Throughout the sequel, we simplify notation by writing:
\begin{equation}\label{eq::muk_rhok}
    \mu(x) = \sum_{k} \mu_k\,1_{X_k}(x), \quad\quad \quad\rho(x) = \sum_{k} \rho_k\,1_{X_k}(x).
\end{equation}

\begin{figure}
    \centering
    \begin{tikzpicture}
    \node at (0,2) {-};
    \node at (-0.3,2.05) {\small 1};
    \node at (4.95,0) {|};
    \node at (4.95,-0.3) {\small 1};
    \node at (0,-0.3) {\small 0};
    \node at (2.5,-0.75) {$f(x)$};
    \draw[->] (0,0) -- (5.2,0);
    \draw[->] (0,0) -- (0,3);
    \draw[dotted] (2,0) -- (2,3);
    \draw[dotted] (2.8,0) -- (2.8,3);
    \draw[dotted] (0,2) -- (5,2);
    
    \draw[->] (2.2,2.5) -- (2,2.5);
    \draw[->] (2.6,2.5) -- (2.8,2.5);
    \node at (2.4,2.45) {$\gamma$};
    \node at (0,3.3) {$\tilde f(x)$};
    
    \node at (2,0) {|};
    \node at (2.8,0) {|};
    
    \draw[thick] plot  coordinates{(0,0) (2,0) (2.8,2) (5,2)};


    \node at (6,1) {-};
    \node at (5.7,1.1) {\small $\frac{1}{4}$};
    \node at (6,2) {-};
    \node at (5.7,2.05) {\small $\frac{1}{2}$};
    \draw[dotted] (6,2) -- (11,2);

    \node at (11,0) {|};
    \node at (11,-0.3) {\small 25};
    \node at (10,0) {|};
    \node at (10,-0.3) {\small 20};
    \node at (9,0) {|};
    \node at (9,-0.3) {\small 15};
    \node at (8,0) {|};
    \node at (8,-0.3) {\small 10};
    \node at (7,0) {|};
    \node at (7,-0.3) {\small 5};

    \node at (6,-0.3) {\small 0};
    \node at (8.5,-0.75) {epochs};
    \draw[->] (6,0) -- (6,3);
    \draw[->] (6,0) -- (11.2,0);
    
    \node at (6,3.3) {$u$};
    
    \draw[thick] plot  coordinates{
(6.0, 0.00) (6.2, 1.07) (6.4, 1.26) (6.6, 1.23) (6.8, 1.23) (7.0, 1.24) (7.2, 1.21) (7.4, 1.25) (7.6, 1.27) (7.8, 1.28) (8.0, 1.27) (8.2, 1.27) (8.4, 1.29) (8.6, 1.28) (8.8, 1.29) (9.0, 1.30) (9.2, 1.29) (9.4, 1.30) (9.6, 1.30) (9.8, 1.30) (10.0, 1.30) (10.2, 1.30) (10.4, 1.30) (10.6, 1.30) (10.8, 1.29)     };
    
    \node at (3,3.5) {\textbf{(a)}};
    \node at (9,3.5) {\textbf{(b)}};
\end{tikzpicture}
    \caption{(a) The $y$-axis displays how the  post-processing decision rule $\tilde f(x)$ looks like as a function of $f(x)$ for each group $X_k\subseteq\mathcal{X}$ that results from the optimization problem in Proposition \ref{prop::unconstraint_opt}. Randomization happens during the transition from $\tilde f(x)=0$ to $\tilde f(x)=1$, where the width of the transition can be controlled using $\gamma$. (b) The value of $u_k$ in one run is plotted against the number of epochs in the stochastic gradient descent method of Section \ref{sect::analysis} for the optimization problem in Proposition \ref{prop::unconstraint_opt}. Here, the learning rate is fixed to $\displaystyle 10^{-1}(K/T)^{1/2}$, where $T$ is the number of steps. 
    }
    \label{fig:fairness_Q}
\end{figure}
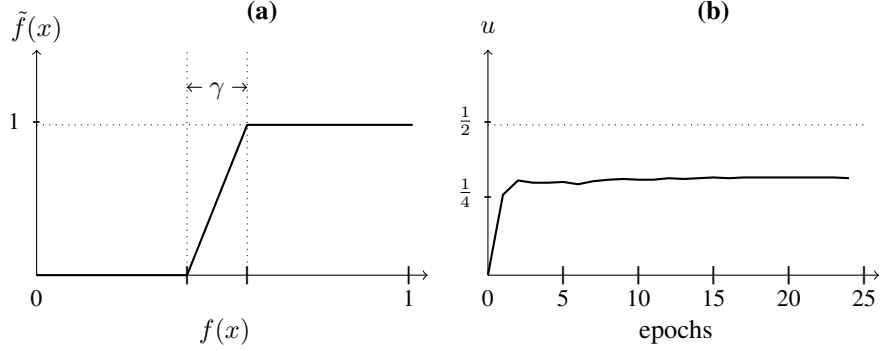

\begin{proposition}\label{prop:thresholding_rule_opt}
    The solution of the optimization problem in Proposition \ref{prop::unconstraint_opt} is the decision rule:
    	\begin{equation}\label{eq::postprocess_eq}
	\tilde f(x) = \begin{cases}
		0,	&\text{if }\; f(x) \le \mu(x)\,(1_S(x)-\rho(x))\\
		1,	&\text{if }\;  f(x) \ge \gamma + \mu(x)\,(1_S(x)-\rho(x))\\
		(1/\gamma)\,(f(x)-\mu(x)\,(1_S(x)-\rho(x))),	&\text{otherwise}.
	\end{cases}
	\end{equation}
\end{proposition}
Proposition \ref{prop:thresholding_rule_opt} shows that the decision rule reduces to a group-specific thresholding rule with randomization near the threshold when $\gamma\to 0^+$, in agreement with the optimal rule stated in Theorem \ref{prop::decision_rules}. The width of the randomization is controlled by $\gamma$ as shown in Figure \ref{fig:fairness_Q}(a).

\subsection{Predictive Equality}
The previous algorithm can be adjusted to control the level of bias according to Definition \ref{def::equal_treatment}. In particular, the equality constraints need to be modified into:
\begin{align}\label{eq::final_lp2}
    \min_{0\le q_i\le 1} \quad\sum_{i=1}^N (\gamma/2)\,q_i^2\,-\,f({x}_i)\,q_i
    &\quad\text{s.t.}\quad\forall X_k\in\mathcal{G}:\; \sum_{i\in X_k} (q_i-\rho) = 0,
\end{align}
where $\rho=\mathbb{E}[f(x)]$ is fixed across all demographic groups. Following a similar proof technique as in Proposition \ref{prop::unconstraint_opt}, we have the following result: 

\begin{proposition}\label{prop::unconstraint_opt_equal_act}
    The optimization problem in (\ref{eq::final_lp2}) is equivalent to minimizing w.r.t $q_i$ and $\mu_k$: 
    \begin{equation}\label{eq::equal_act_opt_form}
        \sum_{X_k\in\mathcal{G}}\sum_{i\in X_k}         \Big(\frac{\gamma}{2}q_i^2 +\rho \mu_k +[f({x}_i)-\gamma q_i-\mu_k]^+\Big),
    \end{equation}
    where $[x]^+ = \max\{0, \,x\}$. In addition, the optimal solution is equivalent to the decision rule:
	\begin{equation}\label{eq::postprocess_eq_equal_act}
	\tilde f(x) = \begin{cases}
		0,&\text{if }\; f(x) \le \mu(x)\\
		1,&\text{if }\;  f(x) \ge \gamma + \mu(x)\\
		(1/\gamma)\,(f(x)-\mu(x)\,),&\text{otherwise}.
	\end{cases}
	\end{equation}
\end{proposition}

\section{Analysis}\label{sect::analysis}

\subsection{Bayes Consistency}
Next, we prove that the classifier learned by the post-processing algorithm converges to the Bayes optimal binary predictor if the original classifier $f$ is itself Bayes consistent. This holds for both conditional statistical parity and predictive equality. The proof of the following theorem is based on Lipschitz continuity of the decision rule when $\gamma>0$ and the robustness-based framework of \cite{xu2012robustness}.

\begin{theorem}\label{prop::excess_risk}
    Let $h^\star = \arg\min_{h\in\mathcal{H}}\,\mathbb{E}[h(\textbf{x})\neq \textbf{y}]$, 
    where $\mathcal{H}$ is the set of all predictors on $\mathcal{X}$ that satisfy fairness according to Definition \ref{def::conditional_convar} (resp. Definition \ref{def::equal_treatment}).
    Let $\tilde h_\gamma:\mathcal{X}\to\{0,1\}$ be  $\tilde h_\gamma(x) = \tilde f(f(x))$, where $\tilde f$ is the optimal solution to  (\ref{eq::theorem_2_equiv_form}) (resp. (\ref{eq::equal_act_opt_form})). If $\tilde f$ is trained  on a freshly sampled data of size $N$, then with a probability of at least $1-\delta$:
    \begin{equation}
        \mathbb{E}[\tilde h_\gamma(\textbf{x})\neq \textbf{y}]\le \mathbb{E}[ h^\star(\textbf{x})\neq \textbf{y}] + \mathbb{E}\,|2\eta(\textbf{x})-1-f(\textbf{x})| + 2\gamma +  \frac{8(2+\frac{1}{\gamma})}{N^{\frac{1}{3}}} +4 \sqrt{\frac{3K+2\log \frac{2}{\delta}}{N}}
    \end{equation}
\end{theorem}
Consequently, if the original classifier is Bayes consistent and we have: $N\to\infty$, $\gamma\to 0^+$ and $\gamma N^{-\frac{1}{3}}\to\infty$, then  $\mathbb{E}[\tilde h_\gamma(\textbf{x})\neq \textbf{y}] \;\xrightarrow{P}\; \mathbb{E}[ h^\star(\textbf{x})\neq \textbf{y}]$. 

\subsection{Running Time Analysis}
The optimization problems for conditional statistical parity in Proposition \ref{prop::unconstraint_opt} and predictive equality in Proposition \ref{prop::unconstraint_opt_equal_act} can be solved using the stochastic gradient descent (SGD) method. As stated earlier, we assume with no loss of generality that $f(x)\in[-1,1]$ since $f(x)$ is assumed to be an estimator to $2\eta(x)-1$ and any thresholding rule over $f(x)$ can be transformed into an equivalent thresholding rule over a monotone increasing function of $f$, such as the hyperbolic tangent. 

Consider the following objective function $F:\mathbb{R}^N\times\mathbb{R}^K\to\mathbb{R}$, which encompasses both conditional statistical parity in (\ref{eq::theorem_2_equiv_form}) and predictive equality in (\ref{eq::equal_act_opt_form}):
\begin{equation}\label{eq:FFp}
    F(q, \mu) = \sum_{i=1}^N         \Big(\frac{\gamma}{2}q_i^2 +b\mu(x_i)+[f({x}_i)-\gamma q_i-\tau(x_i)\mu(x_i)]^+\Big),
\end{equation}
where $b\in\mathbb{R}$, $\mu(x)$ is given by Eq. (\ref{eq::muk_rhok}) and $|\tau(x_i)|\le 1$. Observe that the post-processing rule depends on $\mu(x)$ only so we eliminate the variables $q_i$ by observing that minimizing the objective function in Eq. (\ref{eq:FFp}) is equivalent to minimizing the following functional:
\begin{equation}\label{eq:FF}
    F(\mu) = \sum_{i=1}^N b\mu(x_i) + \xi_\gamma\big(\tau(x_i)\mu(x_i);\,f(x_i)\big),
\end{equation}
\begin{equation}\label{eq::xi}
    \xi_\gamma(z;\,\theta) = \begin{cases} 
        0, &\text{ if } z\ge \theta\\
        \frac{1}{2\gamma} (\theta-z)^2, &\text{ if } \theta-\gamma< z< \theta\\
        \theta-z-\frac{\gamma}{2}, &\text{ if } z\le\theta-\gamma
    \end{cases}
\end{equation}
Note that $\xi_\gamma\in C^1$ (i.e. has a continuous first derivative) and is convex for any $0<\gamma<\infty$. It is a smoothed differentiable approximation to the rectified linear unit (ReLU) \cite{nair2010rectified}. In addition, $|\xi_\gamma'(z)|\le 1$ for all $z\in\mathbb{R}$. The new objective function in Eq (\ref{eq:FF}) contains $K$ optimization variables only, which correspond to the $K$ groups $X_1,\ldots,X_K$.

At each iteration $t$, let $x_t$ be an integer sampled uniformly at random from $\{1,\ldots,N\}$ and write:
\begin{equation}
   \nabla_t = b+
   \frac{\partial }{\partial \mu(x_{i_t})}  \xi_\gamma\big(\tau(x_{i_t})\mu(x_{i_t});\,f(x_{i_t})\big).
\end{equation}
Then, the standard gradient descent method proceeds iteratively by making the updates
for some $\alpha_t$:
\begin{equation}\label{eq:stoch_subgrad_updates}
\mu(x_{i_t}) \leftarrow \mu(x_{i_t}) - \alpha_t \nabla_t
\end{equation}

\begin{proposition}\label{prop::subgradient_udpates}
    Let $\mu^{(0)}=0$ and write  $\mu^{(t)}\in\mathbb{R}^K$ for the value of the optimized function $\mu(x)$ after $t$ stochastic gradient descent updates of the form given in Eq. (\ref{eq:stoch_subgrad_updates}) for some fixed learning rate $\alpha_t=\alpha$. Let $\bar \mu(x) = (1/T)\sum_{t=1}^T \mu^{(t)}(x)$ be the averaged solution. Then, we have:
    \begin{equation}
        \mathbb{E}[F(\bar \mu)] \le F(\mu^\star) +\frac{(1+b)^2\alpha}{2} + \frac{(1+\gamma)^2K}{2T\alpha}, 
    \end{equation}
    where $F:\mathbb{R}^N\times\mathbb{R}^K\to\mathbb{R}$ is the objective function in Eq. (\ref{eq:FF}) and  $\mu^\star(x)$ is its minimizer. In particular,  if $\displaystyle \alpha = ((1+\gamma)/(1+b))\sqrt{K/T}$, then $\mathbb{E}[F(\bar \mu)]-F(\mu^\star) \le
\frac{2(1+\gamma)}{1+b}\sqrt{\frac{K}{T}}$.
\end{proposition}
Hence, the post-processing rule can be learned quite efficiently with negligible computational cost. Figure \ref{fig:fairness_Q}(b) displays the averaged value of $u$ when SGD is applied to the output of the random forests classifier in the Adult dataset to implement statistical parity with respect to the gender attribute (see Section \ref{sect::experiments}). In agreement with Proposition \ref{prop::subgradient_udpates}, convergence is fast. 


\section{Experiments}\label{sect::experiments}
To validate the analysis, we conducted experiments on the Adult dataset from the UCI repository \cite{Dua:2019}, which is one of the most widely used benchmark datasets in the fair machine learning literature (See for instance \cite{zhang2018mitigating,pmlr-v80-agarwal18a,schmidt2018fair,kleindessner2019fair,JMLR:v20:18-262}). The goal of this dataset is to predict if the income of individuals is either above or below \$50K per year. For the purpose of illustration in our experiments, we use gender as a sensitive attribute and let $X_1,\,X_2,\ldots,X_K$ be the racial groups.

The dataset contains 48,842 training instances and 14 attributes that are split into 60\% for training, 20\% for probability calibration and 20\% for testing. Due to space constraint, we only focus on conditional statistical parity.  The original classifiers $f$ we used are: (1) the random forest algorithm in \textbf{scikit-learn} \cite{scikit-learn} in its default settings with maximum depth 10, (2) $k$-NN with $k=10$, and (3) the multi-layer perceptron (MLP)  with Platt's scaling \cite{platt1999probabilistic}. Similar results were obtained with other algorithms, such as logistic regression and support vector machines. The scoring function is $f(x) = 2p(\textbf{y}=1\,|\,\textbf{x}=x)-1$ and $\gamma = 0.01$.

\begin{figure}
    \centering
    \begin{tabular}{ccc}
         Random Forests & $k$-NN &Multilayer Perceptron\\
         \footnotesize {\sc train}: 32\% / 32\% &\footnotesize {\sc train}: 30\% / 30\%  &\footnotesize {\sc train}: 34\% / 35\%\\
         \footnotesize {\sc test}:\;\; 34\% / 35\% &\footnotesize {\sc test}:\;\; 39\% / 38\%  &\footnotesize {\sc test}:\;\; 34\% / 36\%\\
         \includegraphics[width=4.3cm, height=3.5cm]{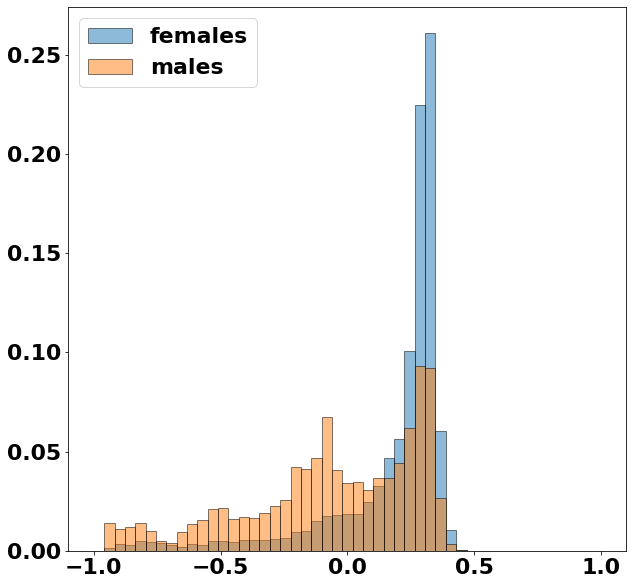}&
         \includegraphics[width=4.3cm, height=3.5cm]{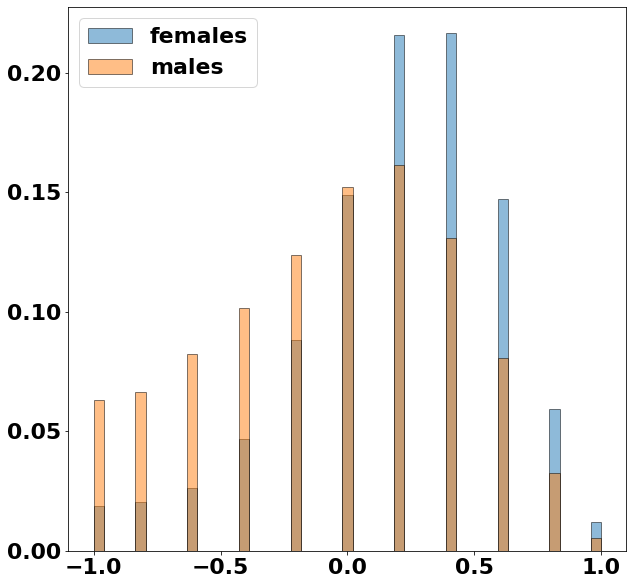}&
         \includegraphics[width=4.3cm, height=3.5cm]{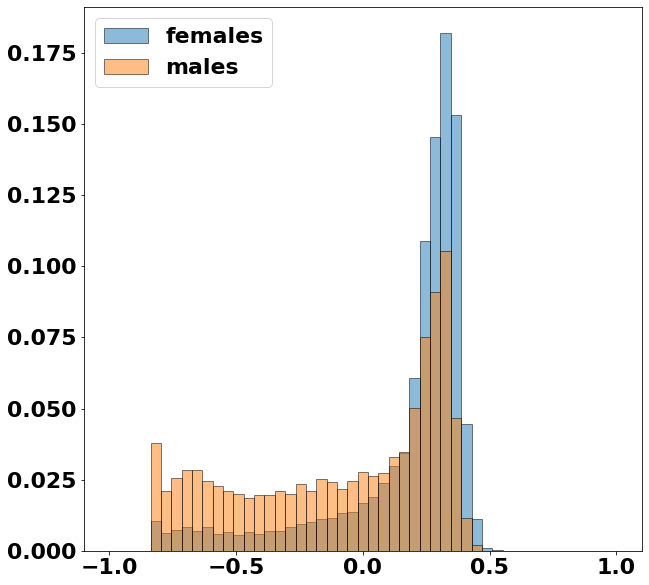}\\
         \includegraphics[width=4.3cm, height=3.8cm]{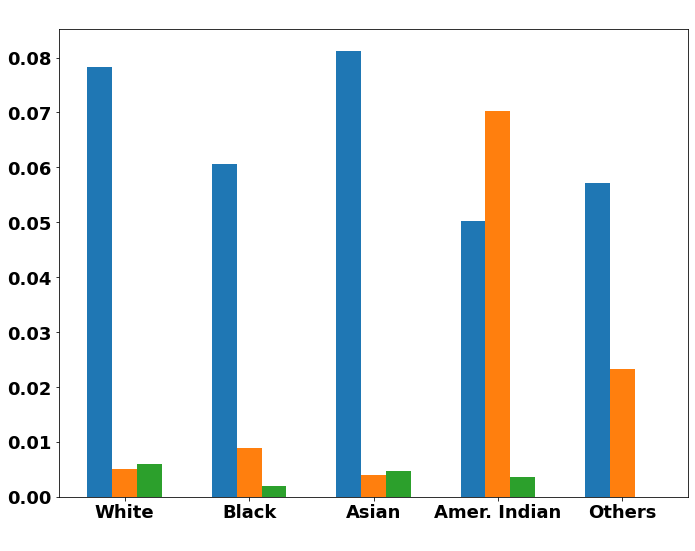}&
         \includegraphics[width=4.3cm, height=3.8cm]{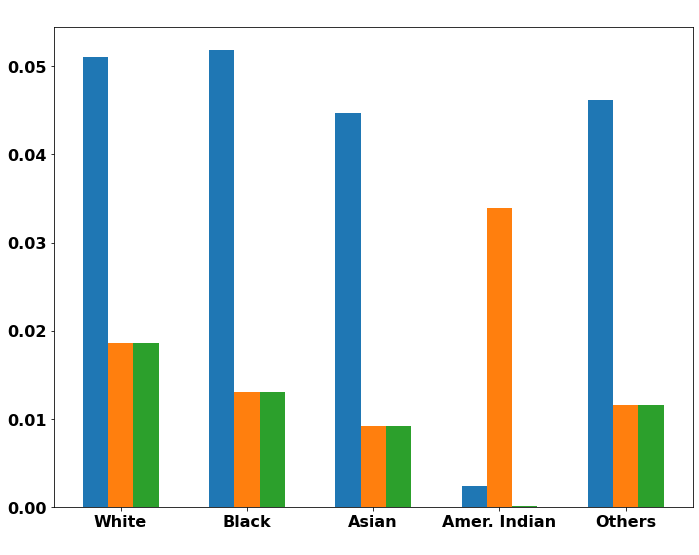}&
         \includegraphics[width=4.3cm, height=3.8cm]{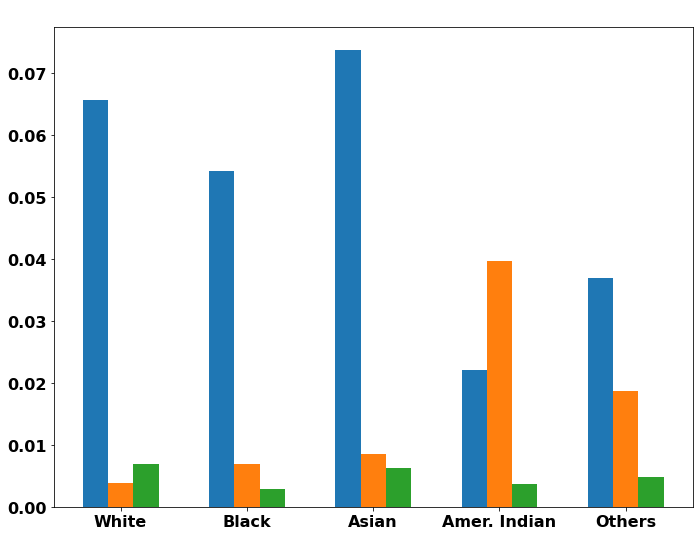}
    \end{tabular}
    \caption{{\sc top} row displays the training/test error rates before/after adjusting the classifier's output using conditional statistical parity. {\sc middle} row displays the histogram of scores $f(x)$ of each classifier for both females and males, demonstrating clear gender bias in all cases. {\sc bottom} row displays the bias before and after applying the proposed algorithm. Blue bars are for the original classifier. Orange bars are for implementing statistical parity with no regard for demographic information. Green bars correspond to conditional statistical parity within every racial group.} 
    \label{fig:hist_scores}
\end{figure}

First, Figure \ref{fig:hist_scores} ({\sc middle}) displays a histogram of the score function $f(x)$ for both females and males and each classifier. As shown in the figure,  all classifiers exhibit unintended bias towards gender. For example, whereas females account for $33\%$ of the data, they comprise $47\%$ of the predicted positive class of the random forests algorithm. One way of removing such bias is to enforce statistical parity across both genders using the proposed algorithm. Doing so would eliminate bias at the expense of a slight increase in test error rate from $34.1\pm 0.3\%$ to $35.0\pm0.3\%$. 

Crucially, however, ensuring statistical parity by itself can lead to different outcomes for different demographic groups. Figure \ref{fig:hist_scores} {(\sc bottom)} displays the bias $|X_k|^{-1} \big|\sum_{i\in X_k} (1_S(i)-\rho_k)\,f(x_i)\big|$ (see Eq. (\ref{eq::final_lp})) before and after adjusting the output.
As shown in the figure, ensuring statistical parity by itself may lead the classifier to discriminate against the sensitive class within one demographic group and compensate for it by favoring the sensitive class in another group. Note in Figure \ref{fig:hist_scores} that controlling statistical parity has \emph{increased} gender discrimination among  American Indians in all three classifiers. Hence, bias needs to be controlled in each demographic group separately. This is one motivation behind the conditional statistical parity in Definition \ref{def::conditional_convar}. The proposed postprocessing algorithm (green bars) is effective at achieving this goal as shown in the figure. Despite the stronger fairness guarantee, its impact on accuracy is negligible as shown in Figure \ref{fig:hist_scores} ({\sc top}).  

\section{Concluding Remarks}
In this paper, we established that the Bayes optimal unbiased classification rule is, in general, a group-wise thresholding rule over the Bayes regressor with a (possible) randomization at the thresholds. This provides a stronger justification to the post-processing approach in fair classification, in which a classifier is learned first before adjusting its output  to remove bias. When the set of group fairness constraints is fixed, Bayes consistent thresholding rules can be learned quite efficiently by solving an unconstrained optimization problem using SGD. We also argued via an impossibility result, that the set of group fairness constraints has to be fixed in advance. The proposed algorithm is provably fast and was shown empirically to have a negligible impact on the accuracy of the original classifier. 

\section*{Statement of Broader Impact}
Machine learning applications are being increasingly adopted to make life-critical decisions with an ever-lasting impact on individual lives, such as for credit lending, medical applications, and criminal justice. Consequently, it is imperative to ensure that such automated decision-making systems abstain from ethical malpractice, including \lq\lq bias." In this work, we show that the post-processing approach in fair classification is near-optimal, whereby a machine learning model is first trained without any fairness constraints and its output is adjusted afterwards to remove bias. It can be applied to any black-box machine learning model, such as deep neural networks, random forests and support vector machines. Hence, existing software implementations for such algorithms can be used without any modification. Moreover,  it can accommodate many fairness criteria that have been previously proposed in the literature, such as equalized odds and statistical parity.

\section*{Acknowledgement}
The author is thankful to Lucas Dixon, Daniel Keysers, Ben Zevenbergen and Olivier Bousquet for the valuable discussions.

\bibliography{fairness}
\bibliographystyle{IEEEtranN}
\newpage
\appendix
\section{Proof of Theorem \ref{prop::decision_rules}}\label{proof:appendixa}
Minimizing the expected misclassification error rate of a classifier $f$ is equivalent to maximizing:
\begin{align*}
    \mathbb{E}[ f(\textbf{x})\cdot \textbf{y}+(1- f(\textbf{x}))\cdot (1-\textbf{y})] &=\mathbb{E}\Big[\mathbb{E}[ f(\textbf{x})\cdot \textbf{y}+(1- f(\textbf{x}))\cdot (1-\textbf{y})]\,\big|\,\textbf{x}\Big]\\
    &=\mathbb{E}\Big[\mathbb{E}[ f(\textbf{x})\cdot (2\eta(\textbf{x})-1)]\,\big|\,\textbf{x}\Big] + \mathbb{E}[1-\eta(\textbf{x})]
\end{align*}
Hence, selecting $f$ that minimizes the misclassification error rate is equivalent to maximizing:
\begin{equation*}
    \mathbb{E}[f(\textbf{x})\cdot (2\eta(\textbf{x})-1)]
\end{equation*}
Instead of maximizing this directly, we consider the regularized form first. Writing $h(x) = 2\eta(x)-1$, the optimization problem is:
\begin{equation*}
    \min_{0\le  f(x)\le 1} \;\; (\gamma/2) \mathbb{E}[f(\textbf{x})^2] - \mathbb{E}[ f(\textbf{x})\cdot h(\textbf{x})] \quad\quad\text{s.t.}\quad\quad \mathbb{E}[w(\textbf{x})\cdot f(\textbf{x})] = b
\end{equation*}
Here, we focused on one subset $X_k$ because the optimization problem decomposes into $K$ separate optimization problems, one for each $X_k$. If there exists a constant $c\in(0,1)$ such that $f(x)=c$ satisfies all the equality constraints, then Slater's condition holds so strong duality holds \cite{boyd2004convex}.

The Lagrangian is:
\begin{equation*}
    (\gamma/2) \mathbb{E}[f(\textbf{x})^2] - \mathbb{E}[ f(\textbf{x})\cdot h(\textbf{x})] + \mu(\mathbb{E}[w(\textbf{x})\cdot f(\textbf{x})]-b) + \mathbb{E}[ \alpha(\textbf{x})(f(\textbf{x})-1)] - \mathbb{E}[\beta(\textbf{x})f(\textbf{x})],
\end{equation*}
where $\alpha(x),\beta(x)\ge 0$ and $\mu\in\mathbb{R}$ are the dual variables. 

Taking the derivative w.r.t. the optimization variable $f(x)$ yields:
\begin{equation}\label{eq::proof_theorem1_fg}
    \gamma f(x)  = h(x)-\mu\,w(x) - \alpha(x) + \beta(x)
\end{equation}
Therefore, the dual problem becomes:
\begin{align*}
    \max_{\alpha(x),\beta(x)\ge 0} -(2
    \gamma)^{-1}\, \mathbb{E}[(h(\textbf{x})-\mu\,w(\textbf{x}) - \alpha(\textbf{x})+\beta(\textbf{x}))^2] -b\mu - \mathbb{E}[\alpha(\textbf{x})]
\end{align*}

We use the substitution in Eq. (\ref{eq::proof_theorem1_fg}) to rewrite it as:
\begin{align*}
    &\min_{\alpha(x),\beta(x)\ge 0} (\gamma/2)\, \mathbb{E}[f(\textbf{x})^2] +b\mu+ \mathbb{E}[\alpha(\textbf{x})]\\
    &\text{s.t.} \forall x\in\mathcal{X}: \gamma f(x)  = h(x)-\mu\,w(x) - \alpha(x) + \beta(x)
\end{align*}
Next, we eliminate the multiplier $\beta(x)$ by replacing the equality constraint with an inequality:
\begin{align*}
    &\min_{\alpha(x)\ge 0} (\gamma/2)\, \mathbb{E}[f(\textbf{x})^2] +b\mu + \mathbb{E}[\alpha(\textbf{x})]\\
    &\text{s.t.} \forall x\in\mathcal{X}: h(x)-\gamma f(x)-\mu\,w(x) - \alpha(x) \le 0
\end{align*}
Finally, since $\alpha(x)\ge 0$ and $\alpha(x)\ge h(x)-\gamma f(x)-\mu w(x)$, the optimal solution is the minimizer to:
\begin{equation*}
    \min_{f:\mathcal{X}\to\mathbb{R}}\;\;(\gamma/2)\mathbb{E}[f(\textbf{x})^2] +b\mu + \mathbb{E}[\max\{0,\,h(\textbf{x})-\gamma f(\textbf{x})-\mu w(\textbf{x})\}]
\end{equation*}

Next, let $\mu^\star$ be the optimal solution of the dual variable $\mu$. Then, the optimization problem over $f$ decomposes into separate problems, one for each $x\in\mathcal{X}$. We have:
\begin{equation*}
    f(x) = \arg\min_{\tau\in\mathbb{R}}\Big\{ (\gamma/2)\tau^2 + [h(x)-\gamma \tau-\mu^\star\,w(x)]^+\Big\}
\end{equation*}

Using the same argument in Appendix \ref{proof:appendix_prop2}, we deduce that $f(x)$ is of the form:
\begin{equation*}
    f(x) = \begin{cases}
        0,  & h(x)-\mu^\star\,w(x)\le 0\\
        1   & h(x)-\mu^\star\,w(x) \ge \gamma\\
        (1/\gamma)\,(h(x)-\mu^\star\,w(x)) &\text{otherwise}
    \end{cases}
\end{equation*}
Finally, the statement of the theorem holds by taking the limit as $\gamma\to 0^+$. 

\section{Proof of Theorem \ref{theorem::impossibility}}
\begin{proof} 
Fix $0<\beta<1$ and consider the subset: 
\begin{align*}
W &= \{x\in \mathcal{X}:\;(\gamma(x)-\bar\gamma)\cdot (f(x)-\beta) > 0 \},
\end{align*}
and its complement $\bar W = \mathcal{X}\setminus W$. Since $f(x)\in\{0,1\}$, the sets $W$ and $\bar W$ are independent of $\beta$ as long as it remains in the open interval $(0,\,1)$. More precisely:
\begin{align*}
W &= \begin{cases} \gamma(x)-\bar \gamma>0 \quad\wedge\quad f(x)=1\\ \gamma(x)-\bar \gamma\le0 \quad\wedge\quad f(x)=0
\end{cases}
\end{align*}

Now, for any set $X\subseteq\mathcal{X}$, let $p_X$ be the projection of the probability measure $p(x)$ on the set $X$ (i.e. $p_X(x) = p(x)/p(X)$). Then, with a simple algebraic manipulation, one has the identity: 
\begin{align}\label{eq::first_identity}
\mathbb{E}_{\textbf{x}\sim p_X}[(\gamma(\textbf{x})-\bar\gamma)\, (f(\textbf{x})-\beta)] = C(\gamma(\textbf{x}), f(\textbf{x});\textbf{x}\in X) + (\mathbb{E}_{\textbf{x}\sim p_X}[\gamma]-\bar\gamma)\cdot (\mathbb{E}_{\textbf{x}\sim p_X}[f]-\beta)
\end{align}
By definition of $W$, we have: 
\begin{align*}\label{eq::inequality_on_w}
\mathbb{E}_{\textbf{x}\sim p_W}&[(\gamma(\textbf{x})-\bar\gamma) (f(\textbf{x})-\beta)] = \mathbb{E}_{\textbf{x}\sim p_W}[|\gamma(\textbf{x})-\bar\gamma| |f(\textbf{x})-\beta|]\ge \min\{\beta, 1-\beta\} \mathbb{E}_{\textbf{x}\sim p_W}|\gamma(\textbf{x})-\bar\gamma| 
\end{align*}

Combining this with Eq. (\ref{eq::first_identity}), we have: 
\begin{equation}\label{eq:c_lower_bound_1}
 C(\gamma(\textbf{x}), f(\textbf{x});\textbf{x}\in W)
\ge \min\{\beta,1-\beta\} \mathbb{E}_{\textbf{x}\sim p_W}|\gamma(\textbf{x})-\bar\gamma| + (\mathbb{E}_{\textbf{x}\sim p_W}[\gamma]-\bar\gamma) (\beta-\mathbb{E}_{\textbf{x}\sim p_W}[f])
\end{equation}
Since the set $W$ does not change when $\beta$ is varied in the open interval $(0,\,1)$, the lower bound holds for any value of $\beta\in(0,1)$. W set:
\begin{equation}\label{eq::beta_star}
\beta = \bar f \doteq \frac{1}{2} \big(\mathbb{E}_{\textbf{x}\sim p_W}f(\textbf{x}) \,+\, \mathbb{E}_{\textbf{x}\sim p_{\bar W}}f(\textbf{x})\big)
\end{equation}
Substituting the last equation into Eq. (\ref{eq:c_lower_bound_1}) gives the lower bound: 
\begin{align}\nonumber
C(\gamma(\textbf{x}), f(\textbf{x});\textbf{x}\in W)
\ge &\min\{\bar f, 1-\bar f\}\cdot \mathbb{E}_{\textbf{x}\sim p_W}|\gamma(\textbf{x})-\bar\gamma|  \\  
&+\frac{1}{2}(\mathbb{E}_{\textbf{x}\sim p_W}[\gamma]-\bar\gamma)\, \big(\mathbb{E}_{\textbf{x}\sim p_W}f(\textbf{x})-\mathbb{E}_{\textbf{x}\sim p_{\bar W}}f(\textbf{x})\big)
\end{align}
Repeating the same analysis for the subset $\bar W$, we arrive at the inequality: 
\begin{align}\nonumber
C(\gamma(\textbf{x}), f(\textbf{x});\textbf{x}\in \bar W)
\le &-\min\{\bar f, 1-\bar f\}\, \mathbb{E}_{\textbf{x}\sim p_{\bar W}}|\gamma(\textbf{x})-\bar\gamma|\\  
&+\frac{1}{2}(\mathbb{E}_{\textbf{x}\sim p_{\bar W}}[\gamma]-\bar\gamma)\, \big(\mathbb{E}_{\textbf{x}\sim p_W}f(\textbf{x})-\mathbb{E}_{\textbf{x}\sim p_{\bar W}}f(\textbf{x})\big)
\end{align}
Writing $\pi(x) = 1_W(x)$, we have by the reverse triangle inequality: 
\begin{align}\label{eq::final_proof}
&\mathbb{E}_{\pi(\textbf{x})}\, \big|\mathcal{C}\big(f(\textbf{\textbf{x}}),\gamma(\textbf{x});\;\pi(\textbf{x})\big)\big| \ge \min\{\bar f, 1-\bar f\}\cdot \mathbb{E}_{\textbf{x}}|\gamma(\textbf{x})-\bar\gamma|
\end{align}
Finally: 
\begin{align*}
2\bar f \ge p(\textbf{x}\in W)\cdot \mathbb{E}_{\textbf{x}\sim p_W}f(\textbf{x})\,+\,p(\textbf{x}\in \bar W)\cdot\mathbb{E}_{\textbf{x}\sim p_{\bar W}}f(\textbf{x}) = \mathbb{E}[f]
\end{align*}
Similarly, we have $2(1-\bar f) \ge 1-\mathbb{E}[f]$. Therefore:
\begin{equation*}
    \min\{\bar f,\,1-\bar f\} \ge \frac{1}{2}\, \min\{\mathbb{E}f, \,1-\mathbb{E}f\}
\end{equation*}
Combining this with Eq. (\ref{eq::final_proof}) establishes the theorem. 
\end{proof}

\section{Proof of Proposition \ref{prop::unconstraint_opt}}
\begin{proof}
First, we observe that the optimization problem in (\ref{eq::final_lp}) decomposes into $|\mathcal{G}|$ separate optimization problems, one for each demographic group $\mathcal{X}_k\in\mathcal{G}$. Hence, we simplify the notation in the proof by considering a single subset only and writing $\rho_k = \rho$ and $\mu_k=\mu$. 

The Lagrangian is:
\begin{align*}
L(q,&\mu,\alpha,\beta) = \sum_i (\gamma/2)\, q_i^2-f(\textbf{x}_i) q_i + \mu\,\sum_i(1_S(i)-\rho)q_i + \sum_i \alpha_i(q_i-1) - \sum_i \beta_i q_i,
\end{align*} 
where $\alpha,\beta\ge 0$. Minimizing this w.r.t. $q_i$ by setting the gradient to zero gives us:
\begin{align*}
L^\star(&\mu,\alpha,\beta)=\sum_i \Big\{ -\frac{1}{2\gamma}\big(\mu(1_S(i)-\rho)+\alpha_i-\beta_i-f(\textbf{x}_i)\big)^2 - \alpha_i \Big\}
\end{align*} 
The original primal optimal solution is recovered by $q_i = (1/\gamma)\,(f(\textbf{x}_i)-\mu(1_S(i)-\rho)-\alpha_i+\beta_i)$. Maximizing this dual objective is equivalent to the unconstrained optimization problem in (\ref{eq::theorem_2_equiv_form}) upon using the dual constraints $\alpha_i,\beta_i\ge 0$. Finally, by Slater's condition \cite{boyd2004convex}, strong duality holds so maximizing the dual is equivalent to minimizing the primal objective. 
\end{proof}

\section{Proof of Proposition \ref{prop:thresholding_rule_opt}}\label{proof:appendix_prop2}
Consider the following the optimization problem for some fixed constants $\gamma>0$ and $\alpha\in\mathbb{R}$:
\begin{equation*}
    \min_{x\in\mathbb{R}}\quad \frac{\gamma x^2}{2} + [\alpha - \gamma x]^+
\end{equation*}
The optimal solution must either be  $\alpha/\gamma$ or is a minimizer to either $\gamma x^2/2+\alpha-\gamma x$ or $\gamma x^2/2$. Hence, the optimal solution must lie in the set $\{0,\,1,\,\,\alpha/\gamma\}$. 

If $\alpha\le 0$, then the optimal solution is $x^\star=0$ since this makes the objective equal to zero and the objective function is always non-negative. 

On the other hand, since $x^\star=1$ is a minimizer to $\gamma x^2/2+\alpha-\gamma x$, the optimal solution is $x^\star=1$ if $\gamma\le \alpha$. Otherwise, it is $\alpha/\gamma$.  


\section{Proof of Theorem \ref{prop::excess_risk}}
We use robustness-based analysis \cite{xu2012robustness}. First, observe that the loss function that we care about is of the form (see Appendix \ref{proof:appendixa}):
\begin{equation*}
    l(u, x_i) = -f(x)\cdot \tilde f(x_i, u),
\end{equation*}
where $f(x):\mathcal{X}\to[-1,+1]$ has a bounded range and $\tilde f$ is of the form shown in Figure \ref{fig:fairness_Q}(a). We will call a function of the form shown in Figure \ref{fig:fairness_Q}(a) a thresholding function with width $\gamma>0$. We note here that the regularization term plays the rule of making the loss class Lipschitz continuous. In general, the decision rule is of the form:
\begin{equation*}
    \tilde f(x) = \begin{cases}
        0,  &\text{ if } f(x)\le \tau_{k,s} \mu_k\\
        1,  &\text{ if } f(x)\ge \gamma + \tau_{k,s}\mu_k\\
        (1/\gamma)\,(f(x)-\tau_{k,s} \mu_k) &\text{ otherwise,}
    \end{cases}
\end{equation*}
for some $|\tau_{k,s}|\le 1$. Note here that $\tau_{k,s}$ depends on the group $X_k$ and the sensitive class $S$ (i.e. all instances $x$ that belong to the same group and sensitive class have the same decision rule). 
In addition, observe that the decision rule depends on $x$ only via $f(x)\in[-1,+1]$. Hence, we write $\textbf{z} = f(\textbf{x})$ and denote the loss by $l(u,\textbf{z}) = \textbf{z}\cdot\tilde f(\textbf{z},\,u)$. Since the thresholds are learned based on a fresh sample of data, the random variables $\textbf{z}_i$ are i.i.d.  Moreover, this loss is $2(1+1/
\gamma)$-Lipschitz continuous within the same group and sensitive class.

Let $\tilde \mu\in\mathbb{R}^K$ be the thresholds produced by the algorithm and let $\tilde{\textbf{h}}_\gamma$ be the resulting decision rule. Using Corollary 5 in \cite{xu2012robustness}, we conclude that with a probability of at least $1-\delta$:
\begin{equation}
   \big| \mathbb{E}_\mathcal{D}[l(\tilde{\textbf{h}}_\gamma,\textbf{x})]-\mathbb{E}_\textbf{s}[l(\tilde{\textbf{h}}_\gamma,\textbf{x})] \big| \le \inf_{R\ge 1} \Big\{\big(\frac{4}{R}(1+\frac{1}{\gamma}\big) +2\sqrt{\frac{2(R+2K)\log 2+2\log \frac{1}{\delta}}{N}} 
   \Big\}
\end{equation}
Here, we used the fact that the observations $f(\textbf{x})$ are bounded in the domain $[-1,1]$ and that we can first partition the domain into groups $X_k$ with/without the sensitive class $S$ ($2K$ subsets) in addition to partitioning the interval $[-1,1]$ into $R$ smaller sub-intervals and using the Lipschitz constant. Choosing $R=N^{\frac{1}{3}}$ and simplifying gives us with a probability of at least $1-\delta$:
\begin{equation*}
    \big| \mathbb{E}_\mathcal{D}[l(\tilde{\textbf{h}}_\gamma,\textbf{x})]-\mathbb{E}_\textbf{s}[l(\tilde{\textbf{h}}_\gamma,\textbf{x})] \big| \le \frac{4(2+\frac{1}{\gamma})}{N^{\frac{1}{3}}} + 2 \sqrt{\frac{3K+2\log \frac{1}{\delta}}{N}}
\end{equation*}

The same bound also applies to the decision rule ${\textbf{h}}^\star_\gamma$ that results from applying optimal threshold $u_\gamma^\star$ with width $\gamma>0$ (here, \lq\lq optimal" is with respect to the underlying distribution) because the $\epsilon$-cover (Definition 1 in \cite{xu2012robustness}) is independent of the choice of the thresholds. By the union bound, we have with a probability of at least $1-\delta$, both of the following inequalities hold:
\begin{align}
    \label{eq::proof_consistency_first_bound}\big| \mathbb{E}_\mathcal{D}[l(\tilde{\textbf{h}}_\gamma,\textbf{x})]-\mathbb{E}_\textbf{s}[l(\tilde{\textbf{h}}_\gamma,\textbf{x})] \big| &\le \frac{4(2+\frac{1}{\gamma})}{N^{\frac{1}{3}}} +2 \sqrt{\frac{3K+2\log \frac{2}{\delta}}{N}}\\
    \label{eq::proof_consistency_second_bound}\big| \mathbb{E}_\mathcal{D}[l( {\textbf{h}}^\star_\gamma,\textbf{x})]-\mathbb{E}_\textbf{s}[l({\textbf{h}}^\star_\gamma,\textbf{x})] \big| &\le \frac{4(2+\frac{1}{\gamma})}{N^{\frac{1}{3}}} +2 \sqrt{\frac{3K+2\log \frac{2}{\delta}}{N}}
\end{align}

In particular:
\begin{align*}
    \mathbb{E}_\mathcal{D}[l(\tilde{\textbf{h}}_\gamma,\textbf{x})] &\le \mathbb{E}_\textbf{s}[l(\tilde{\textbf{h}}_\gamma,\textbf{x})] + \frac{4(2+\frac{1}{\gamma})}{N^{\frac{1}{3}}} +2 \sqrt{\frac{3K+2\log \frac{2}{\delta}}{N}}\\
    &\le \mathbb{E}_\textbf{s}[l( \textbf{h}^\star_\gamma,\textbf{x})] + \gamma + \frac{4(2+\frac{1}{\gamma})}{N^{\frac{1}{3}}} +2 \sqrt{\frac{3K+2\log \frac{2}{\delta}}{N}}\\
    &\le \mathbb{E}_\mathcal{D}[l(\textbf{h}^\star_\gamma,\textbf{x})] +  \gamma + \frac{8(2+\frac{1}{\gamma})}{N^{\frac{1}{3}}} +4 \sqrt{\frac{3K+2\log \frac{2}{\delta}}{N}}
\end{align*}
The first inequality follows from Eq. (\ref{eq::proof_consistency_first_bound}). The second inequality follows from the fact that $\tilde{\textbf{h}}_\gamma$ is an empirical risk minimizer to the regularized loss, where $\mathbb{E}[\tilde f(\textbf{x})^2]\le 1$ since $\tilde f(x)\in[0,1]$. The last inequality follows from Eq. (\ref{eq::proof_consistency_second_bound}). 

Finally, we know that the thresholding rule $\textbf{h}^\star_\gamma$ with width $\gamma>0$ is, by definition, a minimizer to:
\begin{equation*}
    (\gamma/2) \mathbb{E}[h(\textbf{x})^2] - \mathbb{E}[ h(\textbf{x})\cdot f(\textbf{x})]
\end{equation*}
among all possible bounded functions $h:\mathcal{X}\to[0,1]$ subject to the desired fairness constraints. Therefore, we have:
\begin{equation*}
    (\gamma/2) \mathbb{E}[\textbf{h}^\star_\gamma(\textbf{x})^2] - \mathbb{E}[ \textbf{h}^\star_\gamma(\textbf{x})\cdot f(\textbf{x})] \le (\gamma/2) \mathbb{E}[\textbf{h}^\star(\textbf{x})^2] - \mathbb{E}[ \textbf{h}^\star(\textbf{x})\cdot f(\textbf{x})]
\end{equation*}
Hence:
\begin{equation*}
    \mathbb{E}_\mathcal{D}[l(\textbf{h}^\star_\gamma,\textbf{x})] = - \mathbb{E}[ \textbf{h}^\star_\gamma(\textbf{x})\cdot f(\textbf{x})] \le \gamma + \mathbb{E}_\mathcal{D}[l(\textbf{h}^\star,\textbf{x})] 
\end{equation*}
This implies the desired bound:
\begin{equation*}
    \mathbb{E}_\mathcal{D}[l(\tilde{\textbf{h}}_\gamma,\textbf{x})] \le \mathbb{E}_\mathcal{D}[l(\textbf{h}^\star,\textbf{x})] +  2\gamma + \frac{8(2+\frac{1}{\gamma})}{N^{\frac{1}{3}}} +4 \sqrt{\frac{3K+2\log \frac{2}{\delta}}{N}}
\end{equation*}

Therefore, we have consistency if $N\to\infty$, $\gamma\to 0^+$ and $\gamma N^\frac{1}{3}\to\infty$. For example, this holds if $\gamma = O(N^{-\frac{1}{6}})$.

So far, we have assumed that the output of the original classifier coincides with the Bayes regressor. If the original classifier is Bayes consistent, i.e. $\mathbb{E}[|2\eta(\textbf{x})-1-f(\textbf{x})|]\to 0$ as $N\to\infty$, then we have Bayes consistency of the post-processing rule by the triangle inequality. 

\section{Proof of Proposition \ref{prop::subgradient_udpates}}
\begin{proof}
The decision rule by Proposition \ref{prop:thresholding_rule_opt} implies that for all $k\in\{1,\ldots,K\}$, there exists an optimal solution $\mu_k^\star$ that satisfies $ |\mu_k^\star|\le 1+\gamma$. This provides the rationale behind the projection steps. Therefore, we conclude that for any $\mu\in\mathbb{R}^K$, we have the contraction property \cite{boyd2008stochastic}:
\begin{equation}\label{eq::contraction_propety}
   ||\Pi_{[-1-\gamma,\,1+\gamma]}(\mu)-\mu^\star||_2^2 \le || \mu - \mu^\star||_2^2
\end{equation}

Next, since $|\tau(x_i)|\le 1$, we have $||\nabla_t||_2^2\le (1+b)^2$ at all rounds. Finally, following the proof steps of \cite{boyd2008stochastic} and using Eq. (\ref{eq::contraction_propety}), one obtains:
\begin{align*}
    \frac{1}{T}\sum_{t=1}^T \big( \mathbb{E}[F(\mu^{(t)})] - F(\mu^\star)\big) &\le \frac{||\mu^\star||_2^2 + (1+b)^2\sum_{t=1}^T\alpha_t^2}{2T\alpha}\\
    &\le \frac{(1+\gamma)^2K+(1+b)^2 T\alpha^2}{2T\alpha}\\
    &=\frac{(1+b)^2\alpha}{2} + \frac{(1+\gamma)^2K}{2T\alpha}
\end{align*}

The desired result follows by Jensen's inequality since $\frac{1}{T}\sum_{t=1}^T \mathbb{E}[F(\mu^{(t)})] \le \mathbb{E}[F(\bar\mu)]$.
\end{proof}

\end{document}